\documentclass[
    a4paper,
    man,
    floatsintext
]{glossaPX2}

\usepackage[T1]{fontenc}
\usepackage[american]{babel}
\usepackage[style=apa,backend=biber,sorting=nyt,natbib=true]{biblatex}
\NewBibliographyString{unpublished}
\DefineBibliographyStrings{english}{unpublished = {Unpublished}}
\DefineBibliographyStrings{american}{unpublished = {Unpublished}}
\usepackage[font={footnotesize,it}]{caption}
\usepackage{csquotes}
\usepackage{booktabs}
\usepackage{tabularx}
\usepackage{linguex}
\usepackage{cgloss}

\usepackage{amsmath,amssymb,amsfonts,amsthm,mathrsfs}
\makeatletter
\tagsleft@false
\makeatother
\usepackage{graphicx}
\graphicspath{{figures/}}
\usepackage{ragged2e}
\usepackage{hyperref}
\usepackage{enumitem}
\usepackage[title]{appendix}
\usepackage{multirow}
\usepackage{makecell}
\usepackage{array}
\usepackage{adjustbox}
\usepackage{colortbl}
\usepackage{xcolor}
\usepackage{algorithm}
\usepackage{algpseudocode}
\usepackage{listings}
\usepackage{cleveref}
\usepackage{placeins}
\usepackage{float}
\usepackage{subcaption}
\usepackage{rotating}
\usepackage{url}
\usepackage{microtype}

\providecommand{\best}[1]{\cellcolor{green!48}\textbf{#1}}
\providecommand{\second}[1]{\cellcolor{green!18}\textbf{#1}}

\providecommand{\tblwidth}{\textwidth}

\newcolumntype{L}{@{\extracolsep{\fill}}l}
\newcolumntype{C}{@{\extracolsep{\fill}}c}
\newcolumntype{R}{@{\extracolsep{\fill}}r}
\title[Beyond Similarity Matching: Structured Reasoning for Open-Vocabulary Referring Segmentation in 3DGS]{Beyond Similarity Matching: Structured Reasoning for Open-Vocabulary Referring Segmentation in 3DGS}
\author[Wang et al.]
{\spauthor{Yizhao Wang\\
  \institute{School of Computer Science, Henan Institute of Science and Technology}\\
  \small{cswyz@stu.hist.edu.cn, ORCID: 0009-0003-7056-7264}
  }
  \AND
\spauthor{Xinfa Wang\\
  \institute{School of Computer Science, Henan Institute of Science and Technology}\\
  \small{wangxf@hist.edu.cn, ORCID: 0000-0002-6293-5624}
  }
  \AND
\spauthor{Jingbo Wang\\
  \institute{School of Computer Science, Henan Institute of Science and Technology}
  }
  \AND
\spauthor{Ting Li\\
  \institute{School of Computer Science, Henan Institute of Science and Technology}
  }
  \AND
\spauthor{Guantao Zhang\\
  \institute{School of Computer Science, Henan Institute of Science and Technology}
  }
  \AND
\spauthor{Yafeng Han\\
  \institute{School of Computer Science, Henan Institute of Science and Technology}
  }
  \AND
\spauthor{Guohong Gao\\
  \institute{School of Computer Science, Henan Institute of Science and Technology}
  }
  \AND
\spauthor{Yuhe Xia\\
  \institute{School of Computer Science, Henan Institute of Science and Technology}
  }}

\begin{document}
\maketitle

\begin{abstract}
Open-vocabulary referring segmentation in 3D Gaussian Splatting (3DGS) requires a neural model to select Gaussian primitives according to free-form language expressions. Existing 3DGS-based methods usually rely on global text-region similarity, which is weak for queries involving attributes, reference objects, spatial relations, and fine-grained parts. This often causes target-reference confusion, granularity mismatch, part-whole leakage, and relation violations. We propose QAGaussian, a query-adaptive neural reasoning framework for language-guided Gaussian primitive selection. QAGaussian first learns query-conditioned multi-scale Gaussian slots as differentiable candidates whose receptive fields are shaped by the input expression. It then builds a relation-aware slot graph with language-conditioned edge weighting to propagate target-reference, attribute, part-whole, and contextual evidence. A granularity-adaptive router softly combines region-level, object-level, part-level, attribute-aware, and relation-aware mask branches, followed by relation-constrained refinement for spatial, part-whole, attribute, and geometric consistency. QAGaussian is pretrained only on Mosaic3D-5.6M for Gaussian-text alignment and evaluated on independent benchmarks without target-dataset fine-tuning. It achieves 47.2 Avg. mIoU and 63.2 Avg. F1, outperforming the strongest 3DGS referring baseline by 2.7 mIoU points and 2.9 F1 points. It also improves Part-mIoU from 38.6 to 43.4, Rel-mIoU from 44.4 to 50.8, and reduces target-reference confusion from 10.8 to 7.4. These results demonstrate that query-conditioned slot learning, relation-aware graph reasoning, and adaptive routing provide an effective neural modeling strategy for open-vocabulary referring segmentation in 3DGS. The code is available at \url{https://github.com/zqeslwyz/QAGaussian}\mbox{}.
\end{abstract}

\begin{keywords}
  3D Gaussian Splatting; Referring Segmentation; Open-Vocabulary Segmentation; Query-Conditioned Slot Learning; Graph Neural Reasoning; Adaptive Routing
\end{keywords}

\section{Introduction}

Neural vision-language models have made open-vocabulary perception increasingly practical by aligning visual representations with natural language supervision~\citep{radford2021clip,jia2021align,zhai2023siglip,tschannen2025siglip2,wang2024layerwised,li2025confounderprompt}. In 3D scenes, however, language-conditioned perception is not only a recognition problem. A user may ask for ``the red cup to the left of the monitor'', ``the handle of the kettle'', or ``the chair leg near the wall'', where the correct output depends on a target category, attributes, reference objects, spatial relations, and part-whole structure. Such queries require a neural model to learn query-conditioned visual units, reason over relations among candidate units, and adapt the prediction granularity to the expression. This requirement is more demanding than closed-set 3D recognition and motivates a representation-learning formulation for open-vocabulary referring segmentation.

3D Gaussian Splatting (3DGS)~\citep{kerbl2023gaussian} provides an explicit primitive-based scene representation with efficient rendering, making it attractive for semantic feature learning and language-grounded scene understanding. Recent methods attach language or semantic features to Gaussian primitives~\citep{zhou2024feature3dgs,qin2024langsplat,opengaussian2024}, group Gaussians for mask-aware understanding~\citep{ye2024gaussiangrouping,cen2023saga}, and scale Gaussian representation learning with large 3D-language datasets~\citep{ma2025shapesplat,li2025scenesplat,lee2025mosaic3d}. Despite this progress, most open-vocabulary 3DGS segmentation pipelines still reduce a free-form expression to a global text embedding and select regions by text-region similarity. This strategy can work for simple category-name prompts, but it lacks an explicit neural mechanism for decomposing a structured query, constructing query-specific candidates, propagating relation evidence, and selecting an appropriate output granularity.

This limitation leads to four observable failure modes. First, target-reference confusion occurs when the model activates both the referred target and the reference object because they are mentioned in the same sentence. Second, granularity mismatch appears when the expression asks for a part or an attribute-constrained subset but the model predicts a whole object or a broad region. Third, part-whole leakage occurs when a predicted part spills outside its parent object, especially for fine structures defined by PartNet, PartNet-Mobility, and ShapeNetPart~\citep{mo2019partnet,xiang2020partnetmobility,yi2016shapenetpart}. Fourth, relation violation occurs when a mask is semantically plausible but inconsistent with the spatial relation in the query, a problem also observed in language-grounded 3D scene reasoning~\citep{achlioptas2020referit3d,chen2020scanrefer,zhang2023multi3drefer,hovsg2024}. These failure modes indicate that Gaussian-level referring segmentation should be treated as query-conditioned neural reasoning over primitives rather than as static similarity matching.

We propose QAGaussian, a query-adaptive neural framework for open-vocabulary referring segmentation in 3DGS. The core idea is to convert millions of Gaussian primitives into a compact set of differentiable, query-conditioned Gaussian slots, perform relation-aware reasoning over these slots, and dynamically route the prediction to the segmentation branches required by the current expression. Unlike fixed object proposals or direct primitive-text matching, the proposed slot formulation lets the language query shape candidate formation itself. The relation-aware slot graph then parameterizes target-reference, attribute, part, and contextual interactions with language-conditioned edges. Finally, a granularity-adaptive router softly fuses region-level, object-level, part-level, attribute-aware, and relation-aware mask branches, followed by relation-constrained refinement for spatial, part-whole, attribute, and geometric consistency.

QAGaussian is pretrained only on Mosaic3D-5.6M~\citep{lee2025mosaic3d} for Gaussian-text semantic alignment and is evaluated on multiple independent benchmarks without target-dataset fine-tuning. It achieves an Avg. mIoU of 47.2 and an Avg. F1-score of 63.2, improving over the strongest 3DGS referring baseline ReferSplat~\citep{he2025refersplat} by 2.7 mIoU points and 2.9 F1 points under our unified protocol. The gains are more pronounced on complex query types: Part-mIoU improves from 38.6 to 43.4, Rel-mIoU improves from 44.4 to 50.8, and target-reference confusion decreases from 10.8 to 7.4. These results support the claim that query-conditioned slots, relation-aware graph reasoning, and adaptive routing address concrete modeling failures rather than merely adding engineering components. The main contributions are summarized as follows:

\begin{itemize}
    \item We formulate open-vocabulary referring segmentation in 3DGS as query-conditioned Gaussian primitive selection, and introduce a multi-scale Gaussian slot learning mechanism that produces differentiable candidates whose receptive fields are shaped by the input expression.
    \item We develop a relation-aware slot graph with language-conditioned edge weighting to model target-reference, attribute, part-whole, and contextual interactions among Gaussian slots, reducing target-reference confusion and relation violations.
    \item We propose a granularity-adaptive neural routing and consistency learning scheme that softly combines region-level, object-level, part-level, attribute-aware, and relation-aware mask branches and refines the output with spatial, part-whole, attribute, and geometric constraints.
    \item We conduct extensive zero-shot cross-benchmark evaluation, diagnostic analysis, and ablation studies, showing state-of-the-art overall performance and clear gains on part-level, relation-dependent, compositional, and language-robust referring segmentation.
\end{itemize}

\section{Related Work}

\subsection{Open-Vocabulary Segmentation and Gaussian Semantic Learning}

Open-vocabulary segmentation transfers language-aligned visual representations to dense prediction. In 2D vision, CLIP-based dense prediction, mask-level adaptation, unified decoders, and reasoning segmentation models enable segmentation from category names or free-form prompts~\citep{li2022lseg,xu2022groupvit,zhou2022maskclip,liang2023ovseg,zou2023xdecoder,zou2023seem,lai2024lisa}. SAM and language-grounded variants further provide category-agnostic and text-conditioned mask proposals~\citep{kirillov2023sam,li2024semanticsam,liu2024groundingdino,ren2024groundedsam}. These methods provide strong 2D priors, but their masks must be lifted, fused, and reconciled across views before they can support 3D primitive-level prediction.

Open-vocabulary 3D methods extend language-aligned perception to point clouds, neural fields, 3D instances, and semantic maps. DFF and LERF learn language-aligned radiance fields~\citep{kobayashi2022dff,kerr2023lerf}, while OpenScene, ConceptFusion, OpenMask3D, and Open3DIS transfer 2D foundation-model knowledge to 3D scenes~\citep{liu2023openscene,jatavallabhula2023conceptfusion,takmaz2023openmask3d,nguyen2024open3dis}. Recent 3D instruction-tuning and dense-captioning benchmarks further emphasize object-level and scene-level language grounding~\citep{hu20253dbench,yan2025spatialdensecaptioning}. These studies improve open-set recognition, but many of them still treat language mainly as a semantic retrieval signal rather than as a structure that should condition candidate formation and reasoning.

3DGS has recently been extended from real-time rendering to semantic understanding, editing, grouping, and open-vocabulary segmentation~\citep{kerbl2023gaussian}. Feature 3DGS, LangSplat, LEGaussians, and OpenGaussian attach distilled semantic or language features to Gaussian primitives~\citep{zhou2024feature3dgs,qin2024langsplat,shi2023legaussians,opengaussian2024}. Gaussian Grouping, SAGA, GaussianCut, OpenSplat3D, PanoGS, and CAGS explore mask-aware grouping, interactive segmentation, open-vocabulary instance segmentation, panoptic understanding, and context-aware Gaussian reasoning~\citep{ye2024gaussiangrouping,cen2023saga,ye2024gaussiancut,piekenbrinck2025opensplat3d,panogs2025,cags2025}. However, these methods mostly target category-level open-vocabulary segmentation or instance-level recognition. QAGaussian differs by learning query-conditioned Gaussian slots and performing relation-aware neural reasoning for expressions that specify attributes, references, spatial relations, and parts.

\subsection{3D Referring Grounding and Closest 3DGS Baselines}

3D referring expression grounding localizes objects in point clouds or reconstructed scenes according to natural language. ReferIt3D and ScanRefer established object-language matching benchmarks for fine-grained 3D reference resolution~\citep{achlioptas2020referit3d,chen2020scanrefer}. InstanceRefer, 3DVG-Transformer, and TransRefer3D further improve proposal matching and contextual reasoning~\citep{yuan2021instancerefer,zhao2021threedvgtransformer,he2021transrefer3d}. Later methods such as BUTD-DETR, 3D-SPS, MVT, 3D-VisTA, and Multi3DRefer strengthen detection, language-guided point selection, multi-view fusion, 3D vision-language pretraining, and multi-target grounding~\citep{jain2022butddetr,luo2022threedsps,huang2022mvt,zhu20233dvista,zhang2023multi3drefer}. Related visual grounding and referring video object segmentation studies also show the value of semantic decomposition and memory-based target tracking for language-conditioned mask prediction~\citep{wu2025contrastivegrounding,liu2025memoryrvos}.

These grounding methods demonstrate that reference resolution requires more than category recognition, but their outputs are usually bounding boxes, point selections, or object instances. Such outputs are not naturally aligned with 3DGS primitives and are less suitable for part-level or attribute-constrained Gaussian segmentation. ReferSplat is the closest baseline because it explicitly studies referring segmentation in 3DGS and introduces spatially aware language modeling~\citep{he2025refersplat}. OpenGaussian provides open-vocabulary Gaussian-level semantic prediction, while CAGS adds contextual Gaussian reasoning for open-vocabulary scene understanding~\citep{opengaussian2024,cags2025}. Nevertheless, OpenGaussian mainly relies on language-feature similarity, CAGS focuses on context-aware open-vocabulary understanding rather than free-form referring segmentation, and ReferSplat emphasizes spatial-language matching without explicit query-conditioned slot learning or adaptive multi-branch routing. QAGaussian addresses these gaps by making the candidate set, relation graph, and mask granularity all conditioned on the current expression.

Relation-aware 3D scene graphs and open-vocabulary semantic graphs provide another line of structured 3D reasoning. Classic 3D scene graphs model objects, attributes, and spatial relations~\citep{armeni2019threedscenegraph,wald2020threedssg,wu2021scenegraphfusion}, while ConceptGraphs, HOV-SG, Open3DSG, and CLIP-driven scene graph generation connect 3D relations with vision-language models~\citep{gu2023conceptgraphs,hovsg2024,koch2024open3dsg,chen2024clip3dsg}. These graphs are often scene-level and relatively static. Referring segmentation instead requires a compact graph that changes with the expression, since the same object can be a target, reference, contextual distractor, or parent object under different queries. This motivates the query-specific slot graph used in QAGaussian.

\subsection{Slot-Based, Graph-Based, and Adaptive Neural Reasoning}

The proposed framework is also related to object-centric representation learning. Slot Attention learns a set of latent slots that bind visual elements into object-centric representations~\citep{locatello2020object}. This idea is useful for scenes containing multiple entities, but standard slot learning is usually image-level and query-agnostic. In contrast, QAGaussian uses the language expression to modulate slot queries and produces multi-scale Gaussian slots that may correspond to regions, objects, parts, or contextual candidates. This turns slot generation into a query-conditioned primitive selection mechanism rather than a generic object discovery module.

Graph neural networks provide a natural tool for modeling interactions among structured candidates. GCNs, GraphSAGE, and graph attention networks learn representations by propagating information through graph neighborhoods and attention-weighted edges~\citep{kipf2017semisupervised,hamilton2017inductive,velickovic2018graph}. In language-guided 3D segmentation, the graph should not be fixed by geometry alone because the same spatial layout can imply different evidence under different relation words. QAGaussian therefore constructs a relation-aware slot graph whose edge weights depend jointly on spatial descriptors and parsed relation cues, allowing target-reference and part-whole evidence to be interpreted according to the current query.

Adaptive neural computation and routing mechanisms are also relevant. Mixture-of-experts models learn input-dependent combinations of specialized subnetworks~\citep{jacobs1991adaptive,shazeer2017outrageously}, while dynamic routing in capsule networks iteratively assigns lower-level entities to higher-level capsules~\citep{sabour2017dynamic}. QAGaussian follows the general principle of input-conditioned routing, but applies it to Gaussian-level referring segmentation: the router predicts soft weights over region-level, object-level, part-level, attribute-aware, and relation-aware mask branches. This design explicitly links routing decisions to query granularity and slot relations, enabling the model to handle compositional expressions without committing to a single hard query type.

\begin{figure}[t]
  \centering
  \includegraphics[width=\textwidth]{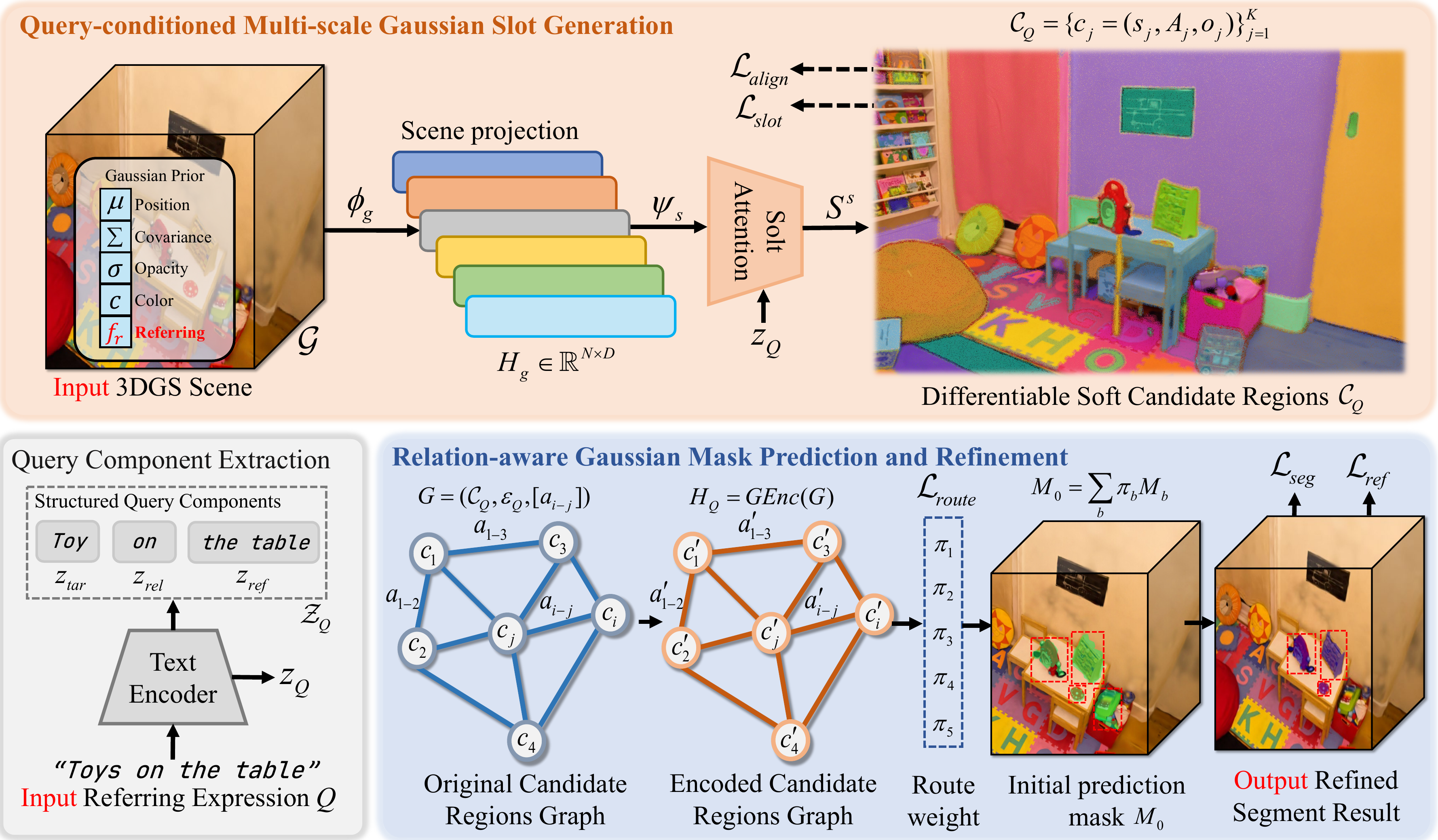}
  \caption{Overall framework of QAGaussian. Given a 3D Gaussian Splatting scene and a free-form referring expression, the model generates query-conditioned multi-scale Gaussian slots, performs relation-aware slot reasoning, adaptively fuses multiple mask prediction branches, and refines the result to obtain the final Gaussian-level referring segmentation mask.}
  \label{fig:qagaussian_framework}
\end{figure}

\section{Methodology}

We formulate open-vocabulary referring segmentation in 3DGS as query-conditioned primitive selection. Given a Gaussian scene $\mathcal{G}=\{g_i\}_{i=1}^{N}$ and a free-form expression $Q$, QAGaussian predicts a probability mask $M\in[0,1]^N$ over Gaussian primitives. As shown in Fig.~\ref{fig:qagaussian_framework}, the framework follows a unified adaptive reasoning pipeline: hierarchical slot generation produces query-specific soft candidates, relation-aware graph reasoning models candidate interactions, and dynamic branch aggregation with geometric refinement yields the final mask. This formulation targets the central ambiguity of natural language queries: the referred entity may correspond to a region, an object, a part, an attribute-constrained subset, or a relation-dependent target.

The design follows two principles. First, all intermediate representations are kept at the Gaussian or slot level, so that the final prediction can be rendered from arbitrary viewpoints without converting back from boxes or point-cloud instances. Second, hard decisions are delayed until the final output stage. Candidate formation, branch selection, and refinement are all represented by soft weights, which makes the pipeline trainable end-to-end and reduces the risk of discarding the correct target early when the expression is ambiguous.

\subsection{Query-Conditioned Hierarchical Slot Generation}

The first question is how to obtain candidate regions from millions of Gaussian primitives without committing to a fixed object proposal set. A referring expression may describe a small part, a complete object, or a broader contextual region, so a single-scale candidate representation is inherently brittle. We therefore use query-conditioned slots to let the language expression decide which Gaussian groups should become candidate units.

QAGaussian first maps the Gaussian scene and the expression into a shared language-aligned feature space. Each Gaussian primitive is represented by its learned appearance feature, opacity-aware geometric attributes, center position, scale, and view-consistent semantic feature distilled during pretraining. These signals are projected by the Gaussian encoder into primitive features $H_G\in\mathbb{R}^{N\times D}$. In parallel, the text encoder produces a global query embedding $z_Q$, and a lightweight parser extracts structured query cues
\begin{equation}
    \mathcal{Z}_Q=\{z_{tar},z_{attr},z_{rel},z_{ref},z_{part}\}.
\end{equation}
These cues represent target, attribute, relation, reference, and part information when they are present in the expression. The parser does not provide query-type labels to the model. Instead, it only decomposes the expression into soft semantic cues, which are later used to condition slot generation, relation weighting, and branch routing.

To capture the hierarchical nature of 3D entities, we pool Gaussian primitives into multi-scale tokens $H_V^s$ and modulate learnable slot queries with the global query context:
\begin{equation}
    H_V^s=\psi_s(H_G),
    \qquad
    Q_c^s=Q_0^s+\mathrm{MLP}_s(z_Q),
    \quad s\in\{1,2,3\}.
\end{equation}
In our implementation, $s=1,2,3$ correspond to fine, middle, and coarse Gaussian token scales. The fine scale preserves local details for small parts and thin structures, the middle scale captures object-level components, and the coarse scale provides larger context for region-level and relation-dependent expressions. The conditioned queries interact with Gaussian tokens through slot attention,
\begin{equation}
    S^s=\mathrm{SlotAttn}_s(Q_c^s,H_V^s),
\end{equation}
where each generated slot represents a soft Gaussian candidate. Unlike a fixed proposal, a slot is not required to correspond to a complete object. It may cover a local part, a complete instance, or a contextual region depending on the query. For slot $j$, its Gaussian assignment and activeness score are predicted as
\begin{equation}
    A_{j,i}=\sigma(s_j^{\top}W_m h_i),
    \qquad
    o_j=\sigma(w_o^{\top}s_j).
\end{equation}
Here, $A_{j,i}$ measures the soft membership of Gaussian primitive $g_i$ to slot $j$, and $o_j$ estimates whether the slot corresponds to a meaningful candidate. Low-activeness slots are not forced to represent physical objects; they serve as inactive capacity that allows the model to use a fixed slot budget across scenes of different complexity. Thus, the model obtains a compact set of query-conditioned candidates whose effective receptive fields are adapted to the expression rather than fixed by a single proposal scale. This step also reduces the computational burden of later reasoning, since graph modeling and routing are performed over slots rather than millions of primitives.

\subsection{Relation-Aware Graph Reasoning and Adaptive Routing}

After candidate generation, the main difficulty shifts from finding plausible regions to selecting the correct one under linguistic constraints. For example, a target and its reference object may both be semantically compatible with the query, but only one satisfies the specified relation or part-whole structure. This motivates explicit relation modeling among slots and a router that adapts the segmentation behavior to the query type.

Slot candidates alone do not resolve expressions involving references, relations, or part-whole structures. We therefore build a lightweight graph over active slots. Nodes are selected according to slot activeness and query relevance, while edges connect spatially close slots, semantically similar slots, and slots that may satisfy relation or containment cues. For slot pair $(i,j)$, the edge descriptor $r_{ij}$ encodes relative position, overlap, containment, scale ratio, and feature similarity. Its query-conditioned edge weight is
\begin{equation}
    a_{ij}=\sigma\left(\mathrm{MLP}_{rel}([r_{ij},z_{rel}])\right),
\end{equation}
which allows the same spatial configuration to be interpreted differently under different relation words. For example, two slots with a similar relative displacement may receive different weights for queries involving ``next to'', ``beneath'', or ``in front of''. This learnable relation weighting avoids relying only on hand-crafted geometric rules and makes the graph more flexible for open-vocabulary expressions. A graph encoder propagates these structural priors and outputs relation-enhanced slot features:
\begin{equation}
    H_Q=\mathrm{GEnc}(S,\mathcal{E}_Q,\mathcal{Z}_Q).
\end{equation}
During message passing, target-like slots receive information from reference-like and context-like slots, while distractor slots are suppressed when they are inconsistent with the query relation. This graph is query-specific: the same scene can produce different effective edges when the expression changes.

The updated slots are passed to a query-adaptive granularity router. The router predicts branch weights $\pi_b$ over region-level, object-level, part-level, attribute-aware, and relation-aware branches. Each branch estimates the contribution of slot $j$ as
\begin{equation}
    p_j^b=\sigma\left(\mathrm{MLP}_b([h_j^Q,z_Q,o_j])\right).
\end{equation}
The branches have different roles. The region-level branch favors broad areas and contextual descriptions, the object-level branch focuses on instance-like targets, the part-level branch emphasizes fine slots and containment cues, the attribute-aware branch strengthens compatibility with color, material, or shape descriptors, and the relation-aware branch relies more on target-reference interactions. The initial Gaussian mask is assembled by weighted slot aggregation:
\begin{equation}
    M_0=\sum_{b}\pi_b\left(\sum_j p_j^b A_j\right).
\end{equation}
This separates what the expression asks for, represented by routing weights, from how the target should be segmented, represented by branch-specific slot aggregation. Because the router outputs a soft distribution rather than a single hard branch, a compositional query can simultaneously use attribute-aware and relation-aware evidence, while a part-level query can still borrow object-level context from its parent object.

\subsection{Relation-Constrained Refinement and Learning Objectives}

The routed mask provides a query-aware prediction, but local Gaussian assignments can still be noisy because 3DGS primitives are optimized for rendering rather than semantic boundaries. In particular, relation-dependent and part-level queries are sensitive to small fragments, leakage to reference objects, and masks that violate containment or spatial consistency. The refinement stage therefore introduces lightweight structural priors without replacing the mask predictor with a heavier decoder.

The initial mask $M_0$ may contain fragmented Gaussians or candidates that violate the query relation. QAGaussian refines it with four lightweight consistency priors: spatial relation consistency, part-whole consistency, attribute compatibility, and Gaussian geometric continuity. These priors form the refinement loss
\begin{equation}
    \mathcal{L}_{ref}
    =
    \lambda_{rel}\mathcal{L}_{rel}
    +\lambda_{part}\mathcal{L}_{part}
    +\lambda_{attr}\mathcal{L}_{attr}
    +\lambda_{geo}\mathcal{L}_{geo},
\end{equation}
where unavailable cues are masked out for expressions that do not contain the corresponding structure. The spatial relation term penalizes masks that do not satisfy the parsed relation with respect to the reference slot. The part-whole term discourages a predicted part from extending outside its parent object candidate. The attribute term improves compatibility between selected Gaussians and attribute cues, and the geometric term suppresses isolated fragments by encouraging local continuity among neighboring primitives. The refinement module updates the mask in a small number of iterations and produces the final soft mask $M_{final}$.

QAGaussian is pretrained end-to-end on Mosaic3D-5.6M without fine-tuning on downstream benchmarks. The training data provides language descriptions and Gaussian-level supervision after unified conversion, which enables Gaussian-language semantic alignment before independent evaluation. Slot-level learning uses Hungarian matching between predicted slots and ground-truth Gaussian masks. The matching cost considers mask overlap, Dice distance, and slot activeness, so that meaningful slots are assigned to annotated masks and redundant slots remain unused. Matched slots are supervised by BCE and Dice mask losses, while unmatched slots are encouraged to remain inactive. The slot objective is
\begin{equation}
    \mathcal{L}_{slot}
    =
    \mathcal{L}_{mask}^{slot}
    +\beta_{act}\mathcal{L}_{act}.
\end{equation}
The router is trained with soft pseudo-labels derived from expression structure and mask geometry, since a single query may require multiple reasoning strategies. For instance, a query containing a part phrase and a spatial relation should not be forced into only one branch. Soft routing supervision allows the model to combine multiple reasoning modes while still learning a preference for the most relevant branch. The full objective is
\begin{equation}
    \mathcal{L}
    =
    \sum_{u\in\Omega}\beta_u\mathcal{L}_u,
    \quad
    \Omega=\{slot,route,seg,align,ref\}.
\end{equation}
Here, $\mathcal{L}_{seg}$ supervises the fused Gaussian mask, $\mathcal{L}_{align}$ aligns positive slot-query pairs and separates negatives, $\mathcal{L}_{route}$ regularizes branch selection, and $\beta_u$ denotes the corresponding loss weight.

During inference, the model takes a reconstructed 3DGS scene and a free-form expression as input. The parser extracts query cues, the slot generator produces multi-scale soft candidates, the graph encoder updates their relation-aware representations, the router fuses branch-specific predictions, and the refinement module outputs $M_{final}$. The model keeps slot assignments and masks soft until the final stage. This avoids early commitment to an incorrect candidate and preserves the information needed by relation-constrained refinement. The discrete segmentation output is obtained by thresholding
\begin{equation}
    \hat{M}=\mathbb{I}(M_{final}>\tau),
\end{equation}
where the threshold and all hyperparameters are fixed after pretraining and shared across independent benchmarks.

\section{Experiments}

\subsection{Experimental Settings}

QAGaussian is pretrained on Mosaic3D-5.6M to learn open-vocabulary Gaussian-language alignment and query-conditioned Gaussian mask prediction. Unless otherwise specified, all hyperparameters are selected on the validation split of Mosaic3D-5.6M and kept fixed across all independent evaluations. We trained and evaluated on eight NVIDIA RTX 2080Ti GPUs.

For visualization, the predicted Gaussian mask can be rendered to arbitrary camera views with the original 3DGS rasterizer. Implementation details are shown in~Table \ref{tab:experimental_settings}.

We compare QAGaussian with five groups of baselines: multi-view 2D projection methods, 3D referring grounding methods, open-vocabulary 3D segmentation and mapping methods, 3DGS open-vocabulary segmentation methods, and 3DGS referring segmentation methods, their predictions are converted to Gaussian primitives using the unified protocol described in Sec. \ref{sec:4.2}.

\begin{table}[t]
\rmfamily
\caption{Experimental settings and main hyperparameters.}
\label{tab:experimental_settings}
\begin{tabular*}{\tblwidth}{@{}LL@{}}
\toprule
Item & Setting \\
\midrule
Pretraining dataset & Mosaic3D-5.6M \\
Hidden dimension $D$ & 256 \\
Gaussian tokenization scales & 3 \\
Slots per scale $(K_1,K_2,K_3)$ & 96 / 64 / 32 \\
Total slots $K$ & 192 \\
Graph neighbors per slot & 16 \\
Slot attention iterations & 3 \\
Graph encoder layers & 2 \\
Refinement iterations & 3 \\
Mask threshold $\tau$ & 0.5 \\
Slot activeness threshold $\tau_o$ & 0.3 \\
Optimizer & AdamW \\
Initial learning rate & $1.0\times10^{-4}$ \\
Weight decay & $1.0\times10^{-2}$ \\
Batch size & 32 \\
Training epochs & 30 \\
Learning rate schedule & Cosine decay \\
Warm-up epochs & 2 \\
\bottomrule
\end{tabular*}
\end{table}

\subsection{Datasets and Evaluation Protocol}
\label{sec:4.2}

For independent evaluation, we use benchmark splits derived from ScanRefer~\citep{chen2020scanrefer}, ReferIt3D~\citep{achlioptas2020referit3d}, Multi3DRefer~\citep{zhang2023multi3drefer}, ReferSplat~\citep{he2025refersplat}, and part-level datasets including PartNet, PartNet-Mobility, and ShapeNetPart~\citep{mo2019partnet,xiang2020partnetmobility,yi2016shapenetpart}. These benchmarks cover object-level, attribute-aware, part-level, relation-dependent, and compositional referring expressions. To ensure fair comparison, we include multi-view 2D projection baselines, 3D referring grounding methods, open-vocabulary 3D segmentation and mapping methods, and 3DGS-native segmentation methods.

Since these benchmarks provide heterogeneous annotations, we convert all supervision and predictions into Gaussian-level masks. For datasets without native 3DGS scenes, we reconstruct a 3DGS representation from posed RGB-D/RGB images and align the original annotation space to the Gaussian coordinate frame using camera poses and rigid registration when necessary. Point-cloud and mesh annotations are transferred to Gaussian primitives by nearest-neighbor or mesh-region assignment based on Gaussian centers. Box annotations are converted by selecting visible Gaussians whose centers fall inside the predicted or ground-truth box. For 2D masks, we render the 3DGS scene from annotated views and accumulate opacity-weighted foreground evidence for each Gaussian, followed by multi-view aggregation and thresholding. The same conversion protocol is applied to baseline predictions, allowing 2D masks, 3D boxes, point-cloud instances, and Gaussian masks to be evaluated under the same Gaussian-level metrics.

We report mIoU and F1-score as the main metrics, and further report Acc@0.25, Acc@0.50, precision, recall, Part-mIoU, Rel-mIoU, target-reference confusion rate, inference time, and memory usage in detailed analyses. Query-type labels are used only for evaluation analysis and are not provided to the model during inference. Custom 3DGS scenes collected by the authors are used only for qualitative visualization and are excluded from training, validation, hyperparameter tuning, and quantitative evaluation.

\subsection{Comparison with State-of-the-Art Methods}
\label{sec:4.3}

\begin{sidewaystable}[p]
\rmfamily
\centering
\caption{Quantitative comparison with state-of-the-art methods on independent benchmarks. Within each method category, darker and lighter green cells indicate the best and second-best results, respectively.}
\label{tab:main_comparison}
\scriptsize
\renewcommand{\arraystretch}{0.90}
\setlength{\tabcolsep}{1.3pt}
\begin{tabular*}{\textheight}{@{\extracolsep{\fill}}lcccccccccccc@{}}
\toprule
\multirow{2}{*}{Method}
& \multicolumn{2}{c}{ScanRefer}
& \multicolumn{2}{c}{ReferIt3D}
& \multicolumn{2}{c}{Multi3DRefer}
& \multicolumn{2}{c}{ReferSplat}
& \multicolumn{2}{c}{Part-level}
& \multicolumn{2}{c}{Avg.} \\
\cmidrule(lr){2-3}\cmidrule(lr){4-5}\cmidrule(lr){6-7}\cmidrule(lr){8-9}\cmidrule(lr){10-11}\cmidrule(l){12-13}
& mIoU & F1 & mIoU & F1 & mIoU & F1 & mIoU & F1 & mIoU & F1 & mIoU & F1 \\
\midrule
\multicolumn{13}{c}{\textit{Multi-view 2D projection methods}} \\
\midrule
G-DINO + SAM + Fusion~\citep{liu2024groundingdino,kirillov2023sam} & 38.4 & 54.0 & 35.2 & 50.8 & 32.0 & 47.0 & 25.4 & 40.5 & 27.0 & 40.8 & 31.6 & 46.6 \\
G-SAM + Fusion~\citep{ren2024groundedsam} & \second{39.8} & \second{55.2} & \second{36.5} & \second{52.0} & \second{34.0} & \second{49.3} & \second{27.6} & \second{42.8} & \second{29.2} & \second{43.0} & \second{33.4} & \second{48.5} \\
CLIPSeg + Fusion~\citep{lueddecke2022clipseg} & 29.5 & 43.8 & 27.4 & 40.5 & 25.6 & 38.0 & 21.2 & 34.5 & 21.0 & 33.2 & 24.9 & 38.0 \\
X-Decoder + Fusion~\citep{zou2023xdecoder} & 38.8 & 54.6 & 35.8 & 51.3 & 33.3 & 48.7 & 28.0 & 43.0 & 29.0 & 42.8 & 33.0 & 48.1 \\
LISA + Fusion~\citep{lai2024lisa} & \best{41.5} & \best{57.0} & \best{38.0} & \best{53.6} & \best{36.2} & \best{51.0} & \best{30.4} & \best{46.0} & \best{31.8} & \best{47.2} & \best{35.6} & \best{50.9} \\
\midrule
\multicolumn{13}{c}{\textit{3D referring grounding methods}} \\
\midrule
ScanRefer~\citep{chen2020scanrefer} & 43.9 & 60.2 & 39.4 & 55.8 & 34.1 & 49.5 & 32.8 & 48.2 & 25.6 & 39.4 & 35.2 & 50.6 \\
InstanceRefer~\citep{yuan2021instancerefer} & 45.2 & 61.3 & 40.5 & 56.5 & 35.0 & 51.1 & 34.3 & 49.8 & 28.5 & 43.8 & 36.7 & 52.5 \\
3DVG-Trans.~\citep{zhao2021threedvgtransformer} & 46.4 & 62.1 & 42.6 & 58.6 & 37.2 & 52.7 & 35.5 & 50.3 & 27.1 & 44.8 & 37.8 & 53.7 \\
TransRefer3D~\citep{he2021transrefer3d} & 47.0 & 63.0 & 43.0 & 59.1 & 38.4 & 54.0 & 36.0 & 51.7 & \second{28.6} & 45.2 & 38.6 & 54.6 \\
BUTD-DETR~\citep{jain2022butddetr} & \best{49.5} & \best{65.0} & \best{44.8} & \best{60.6} & \second{41.6} & \second{57.1} & \second{37.8} & \second{52.8} & 27.3 & 45.0 & \second{40.2} & \second{56.1} \\
3D-SPS~\citep{luo2022threedsps} & 48.1 & 63.8 & 43.8 & 59.7 & 39.5 & 54.9 & 36.9 & 51.4 & 28.4 & \second{46.0} & 39.3 & 55.2 \\
Multi3DRefer~\citep{zhang2023multi3drefer} & \second{49.0} & \second{64.5} & \second{44.1} & \second{60.0} & \best{44.6} & \best{60.2} & \best{38.7} & \best{53.5} & \best{29.2} & \best{46.7} & \best{41.1} & \best{57.0} \\
\midrule
\multicolumn{13}{c}{\textit{Open-vocabulary 3D segmentation and mapping methods}} \\
\midrule
OpenScene~\citep{liu2023openscene} & 46.1 & 62.4 & 42.0 & 58.5 & 37.4 & 53.0 & 30.2 & 45.5 & 29.0 & 41.9 & 36.9 & 52.3 \\
ConceptFusion~\citep{jatavallabhula2023conceptfusion} & 44.8 & 61.0 & 40.3 & 56.8 & 35.9 & 51.6 & 28.8 & 43.9 & 27.8 & 41.3 & 35.5 & 50.9 \\
OpenMask3D~\citep{takmaz2023openmask3d} & \second{48.4} & \second{64.2} & \second{44.9} & \second{60.7} & \second{40.2} & \second{56.2} & \second{33.1} & \second{48.6} & \second{30.8} & \second{47.5} & \second{39.5} & \second{55.4} \\
Open3DIS~\citep{nguyen2024open3dis} & \best{50.1} & \best{65.6} & \best{46.3} & \best{62.3} & \best{42.2} & \best{58.5} & \best{34.4} & \best{50.1} & \best{33.5} & \best{49.8} & \best{41.3} & \best{57.3} \\
\midrule
\multicolumn{13}{c}{\textit{3DGS open-vocabulary segmentation methods}} \\
\midrule
SAGA~\citep{cen2023saga} & 43.2 & 59.6 & 39.9 & 56.0 & 36.6 & 52.0 & 31.5 & 45.6 & 30.1 & 45.4 & 36.3 & 51.7 \\
LangSplat~\citep{qin2024langsplat} & 44.6 & 61.0 & 40.9 & 57.4 & 37.8 & 53.2 & 32.8 & 47.2 & 31.6 & 48.8 & 37.5 & 53.5 \\
OpenGaussian~\citep{opengaussian2024} & 47.5 & 63.4 & 43.6 & 59.7 & 40.8 & 56.1 & 34.7 & 50.2 & 34.9 & 50.4 & 40.3 & 56.0 \\
OpenSplat3D~\citep{piekenbrinck2025opensplat3d} & 49.2 & 65.1 & 45.0 & 61.1 & 42.7 & 58.2 & 36.2 & 52.0 & 36.5 & 52.6 & 41.9 & 57.8 \\
PanoGS~\citep{panogs2025} & \second{50.8} & \second{66.2} & \second{46.7} & \second{62.7} & \second{44.0} & \second{60.1} & \second{37.6} & \second{53.8} & \second{37.4} & \best{53.7} & \second{43.3} & \second{59.3} \\
CAGS~\citep{cags2025} & \best{51.5} & \best{66.8} & \best{47.4} & \best{63.5} & \best{45.2} & \best{61.2} & \best{38.4} & \best{54.6} & \best{37.5} & \second{53.6} & \best{44.0} & \best{59.9} \\
\midrule
\multicolumn{13}{c}{\textit{3DGS referring segmentation methods}} \\
\midrule
ReferSplat~\citep{he2025refersplat} & \best{52.8} & \best{67.6} & \best{48.6} & \best{64.1} & \second{46.9} & \second{62.0} & \second{35.8} & \second{52.6} & \second{38.6} & \second{55.4} & \second{44.5} & \second{60.3} \\
QAGaussian  & \second{52.0} & \second{66.7} & \second{48.0} & \second{63.6} & \best{50.7} & \best{65.2} & \best{41.9} & \best{59.0} & \best{43.4} & \best{61.5} & \best{47.2} & \best{63.2} \\
\bottomrule
\end{tabular*}
\end{sidewaystable}

Table~\ref{tab:main_comparison} and Table~\ref{tab:detailed_average_metrics} reports the quantitative comparison between QAGaussian and 23 representative baselines on multiple independent benchmarks. QAGaussian achieves the best overall performance, with an Avg. mIoU of 47.2 and an Avg. F1-score of 63.2. Compared with ReferSplat~\citep{he2025refersplat}, the strongest 3DGS referring segmentation baseline under our unified protocol, QAGaussian improves Avg. mIoU by 2.7 points and Avg. F1-score by 2.9 points. The gain is more evident on Multi3DRefer, ReferSplat-style evaluation, and part-level evaluation, indicating that QAGaussian is particularly effective for complex referring expressions, spatial relations, and fine-grained parts.

Multi-view 2D projection methods, such as LISA  with Fusion~\citep{lai2024lisa}, benefit from strong 2D vision-language foundation models, but their performance is limited by cross-view inconsistency, occlusion, and target-reference ambiguity during 3D fusion. 3D referring grounding methods, such as BUTD-DETR~\citep{jain2022butddetr} and Multi3DRefer~\citep{zhang2023multi3drefer}, perform well on object-centric queries, but their box-level or instance-level outputs are less suitable for Gaussian primitive-level segmentation, especially for local parts and attribute-constrained regions. Open-vocabulary 3D segmentation methods, such as Open3DIS~\citep{nguyen2024open3dis}, provide stronger open-set recognition, yet they still mainly rely on text-region similarity and therefore struggle with structured expressions requiring relation or part-whole reasoning. This limitation is consistent with recent visual grounding evidence that decomposing language semantics can reduce target-reference confusion~\citep{wu2025contrastivegrounding}.

Among 3DGS-native methods, CAGS~\citep{cags2025} and ReferSplat~\citep{he2025refersplat} are the strongest baselines. ReferSplat performs slightly better than QAGaussian on ScanRefer and ReferIt3D, which are more object-centric. However, QAGaussian achieves clear improvements on Multi3DRefer, ReferSplat-style evaluation, and part-level evaluation. This suggests that strong spatial-language matching is sufficient for many simple object-level expressions, while query-conditioned slot generation, relation-aware graph reasoning, and granularity-adaptive mask prediction are more beneficial for multi-target, relation-dependent, compositional, and part-level queries.

Overall, the results show that QAGaussian improves open-vocabulary referring segmentation in 3D Gaussian Splatting not by uniformly increasing all benchmark scores, but by better modeling the structure of complex referring expressions. This supports the motivation of our framework: Gaussian-level referring segmentation requires semantic alignment, query-adaptive candidate organization, relation reasoning, and granularity-aware mask prediction.

\begin{table}[!b]
\rmfamily
\caption{Average metrics on representative independent benchmarks}
\label{tab:detailed_average_metrics}
\scriptsize
\setlength{\tabcolsep}{5pt}
\begin{tabular*}{\textwidth}{@{\extracolsep{\fill}}lcccccc@{}}
\toprule
Method & mIoU & Acc@0.25 & Acc@0.50 & Precision & Recall & F1 \\
\midrule
G-SAM + Fusion~\citep{ren2024groundedsam} & 33.4 & 53.1 & 28.8 & 47.6 & 49.4 & 48.5 \\
LISA + Fusion~\citep{lai2024lisa} & 35.6 & 55.8 & 31.2 & 50.3 & 51.5 & 50.9 \\
BUTD-DETR~\citep{jain2022butddetr} & 40.2 & 62.0 & 36.9 & 54.8 & 57.5 & 56.1 \\
Multi3DRefer~\citep{zhang2023multi3drefer} & 41.1 & 63.4 & 37.8 & 55.9 & 58.1 & 57.0 \\
OpenMask3D~\citep{takmaz2023openmask3d} & 39.5 & 61.2 & 36.0 & 54.1 & 56.8 & 55.4 \\
Open3DIS~\citep{nguyen2024open3dis} & 41.3 & 64.0 & 38.2 & 56.4 & 58.2 & 57.3 \\
OpenGaussian~\citep{opengaussian2024} & 40.3 & 62.5 & 37.0 & 54.7 & 57.4 & 56.0 \\
CAGS~\citep{cags2025} & 44.0 & 67.5 & 40.9 & 58.6 & 61.3 & 59.9 \\
ReferSplat~\citep{he2025refersplat} & \second{44.5} & \second{68.2} & \second{41.6} & \second{59.0} & \second{61.7} & \second{60.3} \\
QAGaussian & \best{47.2} & \best{72.0} & \best{45.7} & \best{62.4} & \best{64.1} & \best{63.2} \\
\bottomrule
\end{tabular*}
\end{table}

\subsection{Qualitative Comparison with Existing Methods}

\begin{figure}[t]
    \centering
    \includegraphics[width=\textwidth]{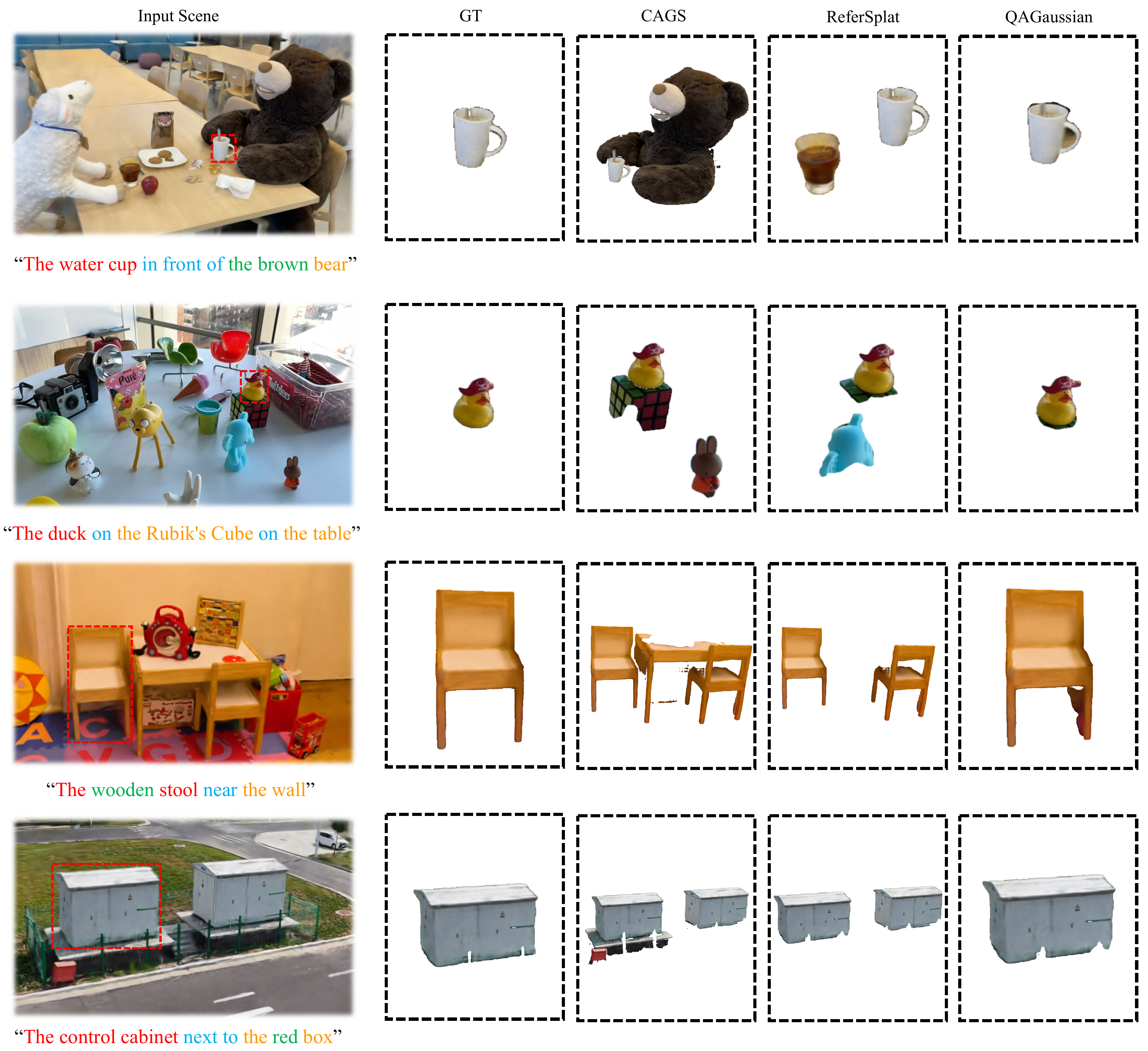}
    \caption{Qualitative comparison with CAGS and ReferSplat on representative 3DGS scenes. Each row shows the input scene, ground-truth mask, and segmentation results of different methods under the same referring expression. The colored query components are automatically identified by the language parser: red denotes the target object, blue denotes the spatial relation, brown denotes the reference object, and green denotes the attribute. QAGaussian produces cleaner Gaussian-level masks and better suppresses reference objects and distractors.}
    \label{fig:qualitative_comparison}
\end{figure}

\label{sec:qualitative_comparison}

Fig.~\ref{fig:qualitative_comparison} shows qualitative comparisons among CAGS, ReferSplat, and QAGaussian on different 3DGS scenes. Each example includes the input scene, ground-truth mask, and segmentation results produced by different methods. The colored words in each query denote the target object, spatial relation, reference object, and attribute, which makes the required language structure explicit.

As shown in the figure, CAGS can often find regions that are semantically related to the query, but it tends to produce coarse or incorrect masks. For example, in the query ``The water cup in front of the brown bear'', CAGS incorrectly segments a large part of the reference object. In the query ``The wooden stool near the wall'', it also includes nearby furniture that does not belong to the target. These results indicate that CAGS has limited ability to distinguish the referred target from reference objects and surrounding distractors.

ReferSplat generally performs better than CAGS and is more suitable for 3DGS scenes. However, it is still affected by similar or nearby objects in complex queries. In the water-cup example, ReferSplat keeps another glass-like object together with the target. In the duck example, it introduces irrelevant objects around the target. In the outdoor control-room example, it also fails to fully suppress the other similar structure. These cases show that direct language-to-Gaussian matching is not sufficient for robust referring segmentation under relation-dependent expressions.

In contrast, QAGaussian produces masks that are closer to the ground truth across these examples. For relation queries, chained spatial queries, and attribute-relation queries, QAGaussian more accurately separates the target from reference objects and distractors. This demonstrates that query-conditioned multi-scale Gaussian slot generation and relation-aware reasoning are effective for open-vocabulary referring segmentation in 3D Gaussian Splatting.

\subsection{Visualization under Different Query Types}
\label{sec:query_type_visualization}
\begin{figure}[t]
    \centering
    \includegraphics[width=\textwidth]{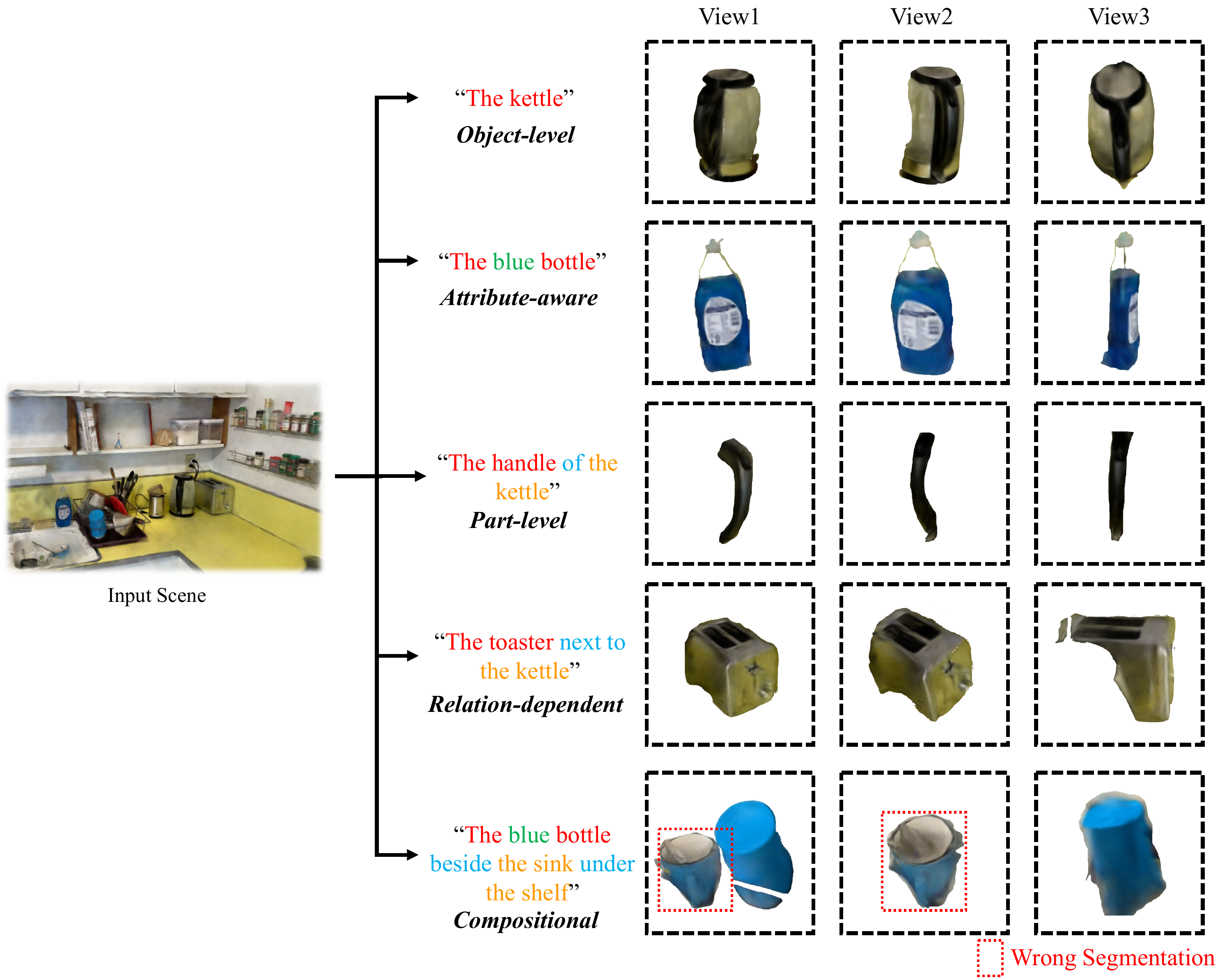}
    \caption{Visualization under different query types within the same complex kitchen 3DGS scene. The input scene is fixed, while the referring expression changes across object-level, attribute-aware, part-level, relation-dependent, and compositional queries. Each row shows Gaussian-level segmentation results from three views. The colored query components are automatically identified by the language parser: red denotes the target object, blue denotes the spatial relation, brown denotes the reference object, and green denotes the attribute.}
     \label{fig:query_type_visualization}
\end{figure}

\label{sec:4.5}
\begin{table}[t]
\rmfamily
\caption{Fine-grained diagnostic metrics on part-level queries from PartNet, PartNet-Mobility, and ShapeNetPart, and relation-dependent queries from ScanRefer, ReferIt3D, Multi3DRefer, and ReferSplat-derived splits. TR-conf. denotes the target-reference confusion rate.}
\label{tab:diagnostic_metrics}
\scriptsize
\setlength{\tabcolsep}{5pt}
\begin{tabular*}{\textwidth}{@{\extracolsep{\fill}}lcccccc@{}}
\toprule
Method & Part-mIoU & Rel-mIoU & TR-conf. $\downarrow$ & Precision & Recall & F1 \\
\midrule
G-SAM + Fusion~\citep{ren2024groundedsam} & 29.2 & 35.0 & 17.2 & 44.6 & 41.5 & 43.0 \\
LISA + Fusion~\citep{lai2024lisa} & 31.8 & 33.4 & 18.9 & 48.6 & 45.9 & 47.2 \\
BUTD-DETR~\citep{jain2022butddetr} & 27.3 & 40.6 & 15.9 & 43.8 & 46.2 & 45.0 \\
Multi3DRefer~\citep{zhang2023multi3drefer} & 29.2 & 41.8 & 15.6 & 45.4 & 48.0 & 46.7 \\
Open3DIS~\citep{nguyen2024open3dis} & 33.5 & 39.6 & 14.8 & 49.2 & 50.4 & 49.8 \\
LangSplat~\citep{qin2024langsplat} & 31.6 & 37.8 & 15.5 & 48.0 & 49.6 & 48.8 \\
OpenGaussian~\citep{opengaussian2024} & 34.9 & 41.5 & 13.9 & 49.5 & 51.3 & 50.4 \\
CAGS~\citep{cags2025} & 37.5 & 43.2 & 12.1 & 52.8 & 54.5 & 53.6 \\
ReferSplat~\citep{he2025refersplat} & \second{38.6} & \second{44.4} & \second{10.8} & \second{54.8} & \second{56.0} & \second{55.4} \\
QAGaussian & \best{43.4} & \best{50.8} & \best{7.4} & \best{60.2} & \best{62.8} & \best{61.5} \\
\bottomrule
\end{tabular*}
\end{table}

Table~\ref{tab:diagnostic_metrics} and Fig.~\ref{fig:query_type_visualization} further analyze the performance under different types of referring expressions. Different from the cross-scene qualitative comparison in Sec.~\ref{sec:qualitative_comparison}, this section uses multiple queries within the same complex kitchen 3DGS scene. This setting removes the influence of scene variation and directly shows whether the model can adapt its segmentation output according to the structure and granularity of the input expression. The query-type labels are used only for analysis and are not provided to the model during inference.

As shown in Table~\ref{tab:diagnostic_metrics}, QAGaussian achieves more stable results on fine-grained and relation-dependent expressions. Compared with ReferSplat, QAGaussian improves Part-mIoU from 38.6 to 43.4 and Rel-mIoU from 44.4 to 50.8, while reducing the target-reference confusion rate from 10.8 to 7.4. This indicates that QAGaussian improves part-level segmentation and reduces confusion between the referred target and reference objects.

Fig.~\ref{fig:query_type_visualization} shows qualitative results for five typical query types in the same scene. For the object-level query ``The kettle'', the model segments the whole kettle consistently across different views. For the attribute-aware query ``The blue bottle'', the model uses the color attribute to select the correct object from multiple container-like objects. For the part-level query ``The handle of the kettle'', the output focuses on the handle instead of the entire kettle, showing that the model can adjust the segmentation granularity according to the query.

For the relation-dependent query ``The toaster next to the kettle'', the model needs to identify both the target and the reference object and use their spatial relation for disambiguation. The result shows that QAGaussian can localize the toaster next to the kettle instead of selecting the kettle itself or other nearby objects. For the compositional query ``The blue bottle beside the sink under the shelf'', the model needs to combine attribute, target, and multiple spatial constraints. Although this query is more challenging and may still be affected by nearby objects in some views, the result remains mainly focused on the target blue bottle.

Overall, Table~\ref{tab:diagnostic_metrics} and Fig.~\ref{fig:query_type_visualization} show that QAGaussian can produce reasonable Gaussian-level masks for different query types. The results are especially clear for part-level, relation-dependent, and compositional expressions, where the model needs to adapt both target selection and segmentation granularity. This supports the effectiveness of query-conditioned multi-scale Gaussian slot generation, relation-aware reasoning, and adaptive routing.

\subsection{Cross-Dataset and Language Robustness Analysis}
\label{sec:cross_dataset_language_robustness}

We further analyze cross-dataset generalization and language robustness. QAGaussian is pretrained only on Mosaic3D-5.6M and is evaluated on independent benchmarks. Therefore, the results reflect whether the learned Gaussian-language alignment can transfer to unseen scenes and annotations.

Table~\ref{tab:pretraining_generalization} reports the effect of different pretraining settings. Without pretraining, the model obtains an Avg. mIoU of 39.6. Using 2D pseudo masks improves the result to 41.2, but the gain is limited because the supervision is not directly aligned with Gaussian primitives. Increasing the Mosaic3D pretraining scale consistently improves cross-dataset performance, and the full Mosaic3D-5.6M setting reaches 47.2 Avg. mIoU without any target benchmark fine-tuning.

\begin{table}[t]
\rmfamily
\caption{Effect of pretraining data on cross-dataset generalization. All settings are evaluated on independent benchmarks without target benchmark fine-tuning.}
\label{tab:pretraining_generalization}
\scriptsize
\setlength{\tabcolsep}{4pt}
\begin{tabular*}{\tblwidth}{@{\extracolsep{\fill}}lccc@{}}
\toprule
Pretraining setting & Avg. mIoU & Acc@0.50 & Avg. F1 \\
\midrule
2D pseudo masks & 41.2 & 37.4 & 56.3 \\
Mosaic3D-1.4M & 43.6 & 40.1 & 58.9 \\
Mosaic3D-2.8M & 45.4 & 42.8 & 61.0 \\
Mosaic3D-5.6M & \textbf{47.2} & \textbf{45.7} & \textbf{63.2} \\
\bottomrule
\end{tabular*}
\end{table}

\begin{figure}
    \centering
    \includegraphics[width=\linewidth]{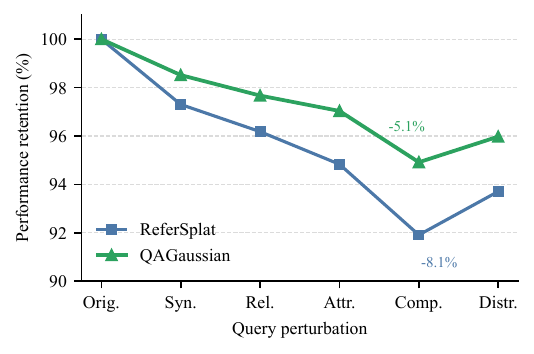}
    \caption{Performance retention under different query perturbations.}
    \label{fig:language_robustness}
\end{figure}

Fig.~\ref{fig:language_robustness} evaluates robustness to language perturbations. We construct perturbed queries by synonym replacement, relation paraphrase, attribute insertion, compositional rewrite, and distractor-rich expression rewriting while keeping the referred target unchanged. The original query is used as the reference point for computing performance retention.

As shown in Fig.~\ref{fig:language_robustness}, QAGaussian retains a larger proportion of its original performance than ReferSplat under all perturbation types. The advantage is especially clear for compositional rewrites and distractor-rich expressions, where direct language-to-Gaussian matching is more likely to lose the target-reference structure. These results suggest that the proposed query-conditioned slot generation and relation-aware reasoning improve both cross-dataset transfer and language robustness.

\subsection{Ablation Study}
\label{sec:ablation_study}

We conduct ablation studies on the independent evaluation benchmarks defined in Sec.~\ref{sec:4.2}, without target benchmark fine-tuning, to evaluate the contribution of each key component in QAGaussian. All variants are trained and evaluated under the same setting as the full model, and only the specified component is removed or replaced. The results are reported in Table~\ref{tab:ablation}.

\begin{table}[!t]
\rmfamily
\caption{Ablation study of key QAGaussian components on the independent evaluation benchmarks. TR-conf. denotes the target-reference confusion rate, where lower values are better.}
\label{tab:ablation}
\small
\setlength{\tabcolsep}{5pt}
\begin{tabular*}{\textwidth}{@{\extracolsep{\fill}}lccccccc@{}}
\toprule
Variant & mIoU & Acc@0.25 & Acc@0.50 & Part-mIoU & Rel-mIoU & TR-conf. $\downarrow$ & F1 \\
\midrule
Full QAGaussian & \textbf{47.2} & \textbf{72.0} & \textbf{45.7} & \textbf{43.4} & \textbf{50.8} & \textbf{7.4} & \textbf{63.2} \\
w/o query-cond. slots & 43.8 & 67.4 & 40.9 & 39.0 & 45.7 & 11.9 & 59.5 \\
w/o multi-scale slots & 44.6 & 68.6 & 42.1 & 40.1 & 46.8 & 10.7 & 60.4 \\
w/o relation graph & 44.9 & 69.0 & 42.4 & 41.0 & 45.9 & 12.6 & 60.8 \\
w/o adaptive router & 45.1 & 69.3 & 42.7 & 40.7 & 47.2 & 10.9 & 61.0 \\
w/o refinement & 45.8 & 70.2 & 43.6 & 41.6 & 48.4 & 9.6 & 61.7 \\
w/o activeness filtering & 46.0 & 70.5 & 43.8 & 42.0 & 48.9 & 9.1 & 61.9 \\
\bottomrule
\end{tabular*}
\end{table}

As shown in Table~\ref{tab:ablation}, the full QAGaussian model achieves the best performance across all metrics. Removing query-conditioned Gaussian slots causes the largest overall degradation, reducing mIoU from 47.2 to 43.8 and F1 from 63.2 to 59.5. This confirms that query-conditioned slot generation is essential for organizing Gaussian primitives into compact and query-relevant soft candidates. Without this component, the model relies more on primitive-level matching and becomes less stable under open-vocabulary referring expressions.

The multi-scale slot design mainly affects fine-grained segmentation. When it is removed, Part-mIoU drops from 43.4 to 40.1, showing that different slot scales are useful for handling part-level and small-region targets. Removing the relation graph leads to the largest drop in Rel-mIoU and the highest TR-conf. value. This indicates that relation-aware reasoning is important for distinguishing the referred target from reference objects and nearby distractors.

The adaptive router and refinement module also provide consistent gains. Without the router, the model cannot dynamically balance object-level, part-level, attribute-aware, and relation-aware branches, leading to weaker performance on both part-level and relation-dependent expressions. Without relation-constrained refinement, the masks become less consistent with spatial, attribute, and geometric cues. Finally, removing activeness filtering causes a smaller but consistent decline, suggesting that suppressing inactive or noisy slots helps stabilize mask prediction. These results demonstrate that the proposed components are complementary and jointly contribute to robust open-vocabulary referring segmentation in 3DGS.

\begin{figure}[!t]
    \centering
    \includegraphics[width=\textwidth]{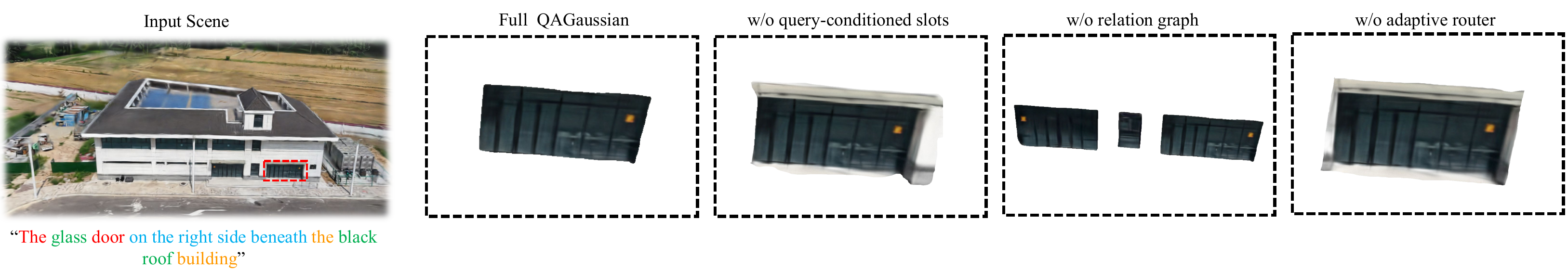}
    \caption{Qualitative ablation example on a relation-dependent region query. The query components are automatically extracted by the language parser and visualized with different colors. The full QAGaussian model accurately segments the referred glass door, while removing query-conditioned slots, relation graph reasoning, or adaptive routing leads to coarse masks, activation of similar glass regions, or leakage to surrounding facade areas.}
    \label{fig:ablation_visualization}
\end{figure}

Fig.~\ref{fig:ablation_visualization} further provides a qualitative ablation example on the query ``The right glass door beneath the black roof''. The full model accurately localizes the referred glass door and suppresses surrounding windows, facade regions, and roof structures. This indicates that QAGaussian can combine the target description, positional cue, and reference object to produce a compact Gaussian-level mask.

When query-conditioned slots are removed, the predicted mask expands to nearby facade and roof-adjacent regions, suggesting that the model fails to form compact candidates conditioned on the current expression. Without relation graph reasoning, the prediction is split into several visually similar glass regions, which shows that relation-aware slot interactions are important for spatial disambiguation. Without the adaptive router, the output becomes coarser and includes more surrounding building surfaces, indicating that dynamic branch selection helps match the required segmentation granularity. This qualitative evidence is consistent with the quantitative trends in Table~\ref{tab:ablation}.

\subsection{Efficiency Analysis}
\label{sec:mechanism_efficiency}

We further analyze the computational behavior of QAGaussian. Although the proposed framework introduces query-conditioned slot generation, relation-aware graph reasoning, and relation-constrained refinement, these operations are performed on a compact set of Gaussian slots rather than all Gaussian primitives. Therefore, the additional cost mainly comes from slot construction, graph message passing, and a small number of refinement iterations.

\begin{table}[b]
\rmfamily
\caption{Efficiency comparison on Gaussian scenes of different scales. Preprocess and query time are reported for small, medium, and large scenes.}
\label{tab:efficiency}
\scriptsize
\setlength{\tabcolsep}{4pt}
\begin{tabular*}{\textwidth}{@{\extracolsep{\fill}}lcccccccccc@{}}
\toprule
Method
& \multicolumn{3}{c}{Preprocess time}
& \multicolumn{3}{c}{Query time}
& \multicolumn{3}{c}{Memory}
& Avg. mIoU \\
\cmidrule(lr){2-4}\cmidrule(lr){5-7}\cmidrule(lr){8-10}
& Small & Medium & Large
& Small & Medium & Large
& Small & Medium & Large
&  \\
\midrule
OpenGaussian~\citep{opengaussian2024}
& \best{6.2m} & \best{8.6m} & \best{13.8m}
& \best{0.28s} & \best{0.43s} & \best{0.68s}
& \best{5.8GB} & \best{7.8GB} & \best{10.6GB}
& 40.3 \\
CAGS~\citep{cags2025}
& \second{6.8m} & \second{9.4m} & \second{15.1m}
& \second{0.36s} & \second{0.61s} & \second{0.92s}
& \second{6.3GB} & \second{8.6GB} & \second{11.8GB}
& 44.0 \\
ReferSplat~\citep{he2025refersplat}
& 7.1m & 9.8m & 15.8m
& 0.40s & 0.66s & 1.05s
& 6.8GB & 9.1GB & 12.4GB
& \second{44.5} \\
QAGaussian
& 7.4m & 10.3m & 16.6m
& 0.48s & 0.82s & 1.28s
& 7.5GB & 9.8GB & 13.6GB
& \best{47.2} \\
\bottomrule
\end{tabular*}
\end{table}

Table~\ref{tab:efficiency} reports the runtime and memory usage on Gaussian scenes of different scales. OpenGaussian is the fastest method because it mainly relies on direct open-vocabulary similarity matching, and CAGS also keeps relatively low computational cost. ReferSplat introduces spatially aware referring segmentation and therefore requires slightly more computation. QAGaussian is slower than these baselines, but the gap is moderate: on medium-scale scenes, the query time increases from 0.66s for ReferSplat to 0.82s for QAGaussian, while Avg. mIoU improves from 44.5 to 47.2.

The additional cost is consistent with the design of QAGaussian. Query-conditioned multi-scale Gaussian slot generation avoids exhaustive primitive-level reasoning, but it still needs to build soft slot assignments for each expression. The relation-aware graph encoder then updates slot representations using target-reference and spatial cues, and the refinement module improves mask consistency through several lightweight iterations. These steps increase runtime and memory slightly, but they lead to clearer gains on relation-dependent, compositional, and part-level expressions, as shown in Tables~\ref{tab:diagnostic_metrics} and~\ref{tab:ablation}.

Overall, QAGaussian provides a favorable accuracy-efficiency trade-off. It is not designed to be the fastest similarity-matching method; instead, it adds a compact reasoning layer over Gaussian slots to improve open-vocabulary referring segmentation. The efficiency results show that this structured reasoning brings a 2.7-point Avg. mIoU improvement over ReferSplat with acceptable additional query time and memory usage.

\section{Discussion}
\label{sec:discussion}

\textbf{Effectiveness of query-conditioned Gaussian slots.}
The main improvement of QAGaussian comes from organizing Gaussian primitives in a query-conditioned manner rather than relying on a fixed candidate pool or direct text-region matching. In open-vocabulary referring segmentation, the same 3DGS scene may require different segmentation granularity under different expressions. For example, a query may refer to a whole object, a local part, an attribute-constrained object, or a target determined by its relation to another object. Fixed candidates are often insufficient for such flexible language structures, especially when spatial perception and contextual 3D descriptions are required~\citep{yan2025spatialdensecaptioning}. By generating multi-scale Gaussian slots conditioned on the input expression, QAGaussian can form compact soft candidates that are more relevant to the current query. This design explains why the model performs better on part-level, relation-dependent, and compositional expressions, and why removing query-conditioned slots leads to the largest degradation in the ablation study.

\textbf{Role of relation-aware reasoning.}
A central challenge in 3D referring segmentation is not only recognizing the target category, but also distinguishing the target from reference objects and nearby distractors. Many expressions contain explicit spatial relations, such as ``next to'', ``beneath'', or ``in front of'', where the reference object should guide target localization but should not be included in the final mask. Direct language-to-Gaussian similarity matching tends to confuse the referred target with the reference object or other semantically similar regions. The relation-aware slot graph in QAGaussian provides a compact structure for modeling interactions among target, reference, attribute, part, and contextual slots. This helps the model suppress reference leakage and reduce target-reference confusion, especially in relation-dependent queries. The diagnostic metrics and qualitative ablation results show that relation reasoning is particularly important for improving Rel-mIoU and reducing TR-conf.

\textbf{Generalization without target benchmark fine-tuning.}
QAGaussian is pretrained only on Mosaic3D-5.6M for Gaussian-language semantic alignment and is evaluated on independent benchmarks. This setting is important because open-vocabulary 3D scene understanding should not depend on adapting the model to every downstream dataset. The pretraining analysis shows that larger-scale Mosaic3D pretraining improves cross-dataset performance, suggesting that Gaussian-level language alignment can transfer across different scene types, annotation formats, and query distributions. Compared with using frozen 2D vision-language features or 2D pseudo masks alone, Mosaic3D pretraining provides supervision that is more directly aligned with Gaussian primitives. This supports the use of large-scale Gaussian-language pretraining as a practical foundation for open-vocabulary 3DGS segmentation.

\textbf{Efficiency and practical trade-off.}
QAGaussian is not the fastest method among the compared approaches, since it introduces query-conditioned slot generation, relation-aware graph reasoning, adaptive routing, and relation-constrained refinement. However, these operations are performed over a compact set of slots rather than all Gaussian primitives, which keeps the additional cost moderate. The efficiency analysis shows that QAGaussian requires slightly higher query time and memory than simpler similarity-matching methods, but provides a clear improvement in segmentation accuracy, especially for complex referring expressions. This trade-off is reasonable for applications such as embodied interaction, robotic manipulation, and augmented reality, where correctly identifying the referred 3D target is often more important than minimizing a small amount of additional inference time.

\textbf{Limitations and future work.}
Despite its effectiveness, QAGaussian still has several limitations. First, the method depends on the quality of the reconstructed 3DGS scene. If the Gaussian representation is noisy, incomplete, or inaccurate due to occlusion, transparent objects, reflective surfaces, or insufficient views, the predicted mask may also be degraded. Second, very small parts and thin structures remain challenging, because they require both high-quality geometry and fine-grained semantic alignment. Third, although the relation-aware slot graph improves spatial reasoning, highly complex multi-hop expressions, ambiguous references, and relations such as ``between'', ``inside'', or ``aligned with'' may still be difficult. Future work may explore uncertainty-aware slot selection, stronger 3D scene graph priors, interactive correction, and extension to dynamic or 4D Gaussian scenes. These directions could further improve the robustness and applicability of open-vocabulary referring segmentation in 3D Gaussian Splatting.

\section{Conclusion}
\label{sec:conclusion}

This paper presented QAGaussian, a query-adaptive framework for open-vocabulary referring segmentation in 3D Gaussian Splatting. Unlike existing methods that mainly rely on direct text-region similarity matching, QAGaussian models the internal structure of free-form referring expressions and performs Gaussian-level segmentation through query-conditioned multi-scale slot generation, relation-aware slot graph reasoning, adaptive granularity routing, and relation-constrained mask refinement. By organizing Gaussian primitives into differentiable query-conditioned slots, the proposed framework can adapt its target selection and segmentation granularity to object-level, attribute-aware, part-level, relation-dependent, and compositional expressions.

QAGaussian is pretrained on Mosaic3D-5.6M for Gaussian-language semantic alignment and evaluated on independent benchmarks. Extensive experiments show that QAGaussian achieves state-of-the-art overall performance and provides clear gains on complex referring expressions, part-level segmentation, relation-dependent queries, language robustness, and cross-dataset generalization. Ablation studies further verify the complementary contributions of query-conditioned slots, multi-scale modeling, relation-aware reasoning, adaptive routing, and refinement. Although the method introduces moderate additional computation compared with simpler similarity-matching baselines, it offers a favorable accuracy-efficiency trade-off for language-guided 3DGS scene understanding. Future work will explore stronger multi-hop relation modeling, uncertainty-aware slot selection, interactive correction, and extension to dynamic or 4D Gaussian scenes.

\section*{Acknowledgments}
This research was supported by the Henan Province Key Research and Development Program (Grant No. 251111210800, Grant No. 261111210600).

\section*{Declaration of Competing Interest}
The authors declare that they have no known competing financial interests or personal relationships that could have appeared to influence the work reported in this paper.

\section*{Data Availability}
Data will be made available on request.

\nocite{*}




\printbibliography

\end{document}